\documentclass[times,twocolumn,final]{elsarticle}
\usepackage{prletters}
\usepackage{framed,multirow}

\usepackage{amssymb}
\usepackage{latexsym}
\usepackage{amsmath}

\usepackage{url}
\usepackage{xcolor}

\usepackage{amsmath}
\usepackage{algpseudocode,algorithm,algorithmicx}
\usepackage{balance}
\usepackage{booktabs}

\newcommand{\ru}{\rule{0mm}{2.4mm}}
\newcommand{\red}   {}
\renewcommand{\snm}   {}

\newcommand{\winner}{{(grip)}}
\newcommand{\G}     {{\cal G}}
\renewcommand{\L}   {{\cal L}}
\newcommand{\SSIM}  {{\rm SSIM}}

\newcommand{\NA} {$\mathcal{N}$Attack}
\newcommand{\U} {$\uparrow$}
\newcommand{\D} {$\downarrow$}

\journal{}

\begin{document}

\thispagestyle{empty}

\begin{frontmatter}

\title{Perceptual quality-preserving black-box attack against deep learning image classifiers}

\author[1]{Diego \snm{Gragnaniello}\corref{cor1}}
\cortext[cor1]{Corresponding author:
}
\ead{diego.gragnaniello@unina.it}
\author[1]{Francesco \snm{Marra}}
\author[2]{Luisa \snm{Verdoliva}}
\author[1]{Giovanni \snm{Poggi}}

\address[1]{Department of Electrical Engineering and Information Technology, University Federico II of Naples, via Claudio, 21, 80125 Naples, Italy}
\address[2]{Department of Industrial Engineering, University Federico II of Naples, via Claudio, 21, 80125 Naples, Italy}


\begin{abstract}
Deep neural networks provide unprecedented performance in all image classification problems, including biometric recognition systems, key elements in all smart city environments.
Recent studies, however, have shown their vulnerability to adversarial attacks, spawning an intense research effort in this field.
With the aim of building better systems, new countermeasures and stronger attacks are proposed by the day.
On the attacker's side,
there is growing interest for the realistic black-box scenario, in which the user has no access to the neural network parameters.
The problem is to design efficient attacks which mislead the neural network without compromising image quality.
In this work, we propose to perform the black-box attack along a high-saliency and low-distortion path, so as to improve both the attack efficiency and the {\em perceptual} quality of the adversarial image.
Numerical experiments on real-world systems prove the effectiveness of the proposed approach both on benchmark tasks and actual biometric applications.
\end{abstract}


\end{frontmatter}


\section{Introduction}
\label{sec:introduction}

Deep Neural Networks (DNNs) are by now widespread in industry and society as a whole,
finding application in uncountable fields, from the movie industry, to autonomous driving, humanoid robots, video surveillance, and so on.
Well trained DNNs largely outperform conventional systems, and can compete with human experts on a large variety of tasks.
In particular,
there has been a revolution in all vision-related tasks, which now rely almost exclusively on deep-learning solutions,
starting from the 2012 seminal work of Krizhevsky {\it et al.} \cite{Krizhevsky2012}
where state-of-the-art image classification performance was achieved with a convolutional neural network (CNN).

\begin{figure}
	\centering
	\includegraphics[width=0.75\linewidth]{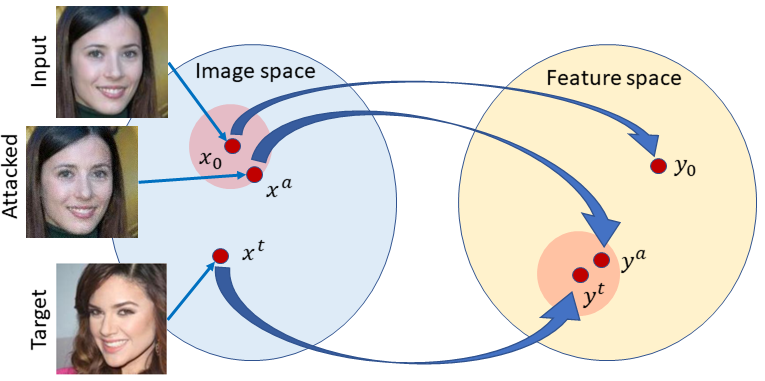}
	\caption{Attack scenario. The attacker adds adversarial noise $W$ to the original image $x_0$ to generate $x^a=x+W$.
    This attacked image should be similar to $x_0$ but such that the associated CNN output $y^a$ is close to the target $y^t$.}
	\label{fig:mapping}
\end{figure}

Recent studies \cite{Szegedy2014}, however, have exposed some alarming weaknesses of DNNs.
By injecting suitable {\em adversarial noise} on a given image,
a malicious attacker can mislead a DNN into deciding for a wrong class, and even force it to output a {\em desired} wrong class, selected in advance by the attacker, a scenario described in Fig.\ref{fig:mapping}.
What is worse, such attacks are extremely simple to perform.
By exploiting backpropagation, one can compute the gradient of the loss with respect to the input image,
and build effective adversarial samples by gradient ascent/descent methods.
Large loss variations can be induced by small changes in the image, ensuring that adversarial samples keep a good perceptual quality.


Following \cite{Szegedy2014}, several attacks have been proposed\footnote{As well as defenses, not mentioned here for brevity.},
mostly gradient-descent methods with the gradient estimated through backpropagation,
from the early Fast Gradient Sign Method (FGSM) \cite{Goodfellow2015}, and its iterative version (I-FGSM) \cite{Kurakin2016}, to more recent and sophisticated methods \cite{Carlini2017, Madry2017, Cisse2017}.
All such methods, however, require perfect knowledge of the network architecture and weights,
a {\em white box} scenario which is hardly encountered in real-world applications.
The focus is therefore shifting towards {\em black-box} attacks,
where nothing is known in advance about the network structure, its weights, or the dataset used for training.
In this scenario, the attacker can query the network at will and observe the outcome.
This latter can be just a hard label, the distribution of probabilities across the classes (confidence levels), or even a feature vector.

{\red
There are many ways to perform black-box attacks to classifiers.
A popular approach is to train a surrogate network to mimic the behavior of the target network \cite{Carlini2017, Papernot2016a, Papernot2017practical}.
The attack is performed on the surrogate and then transferred to the target.
However, this calls for full knowledge of the target training set, a precious information rarely available in practice.
A more viable alternative, followed here, is to use again gradient-descent \cite{Chen2017zoo,Narodytska2017,Bhagoji2017exploring,Tu2019,Ilyas2018},
with the gradient now estimated by means of suitable queries to the network.
}

Black-box adversarial attacks should satisfy three main requirements at the same time:
being {\it i)} effective, {\it ii)} fast, and {\it iii)} inconspicuous.
The first requirement needs no comment, as it is the primary goal of the attack.
Efficiency (low number of queries) is also necessary to prevent the attack from becoming exceedingly slow.
As for the third requirement, it is important in all systems with human-in-the-loop, such as semiautomatic biometric recognition systems.
Unnatural, low quality images may be readily identified by dedicated statistical tests and sent to visual inspection,
to be eventually easily detected.

To meet all these requirements, we propose
a novel, perceptual quality-preserving (PQP), black-box attack, where quality and effectiveness are ensured in advance by a judicious choice of the perturbation path.
Let us briefly summarize the major features and innovative contributions of our proposal:
{\it   i)} adversarial noise is injected only in ``safe'' regions where it has low impact on image quality and high impact on decisions;
{\it  ii)} safe regions are identified based on the local gradient of a perceptual quality measure, the structural similarity index (SSIM) \cite{Wang2004};
{\it iii)} the SSIM gradient is computed with negligible cost by means of a simple dedicated CNN;
{\it  iv)} all perturbations and queries are 8-bit integer, to successfully attack systems that accept only popular integer formats, such as PNG or JPG.

We assess performance both on widespread benchmark datasets and on a realistic biometric application, including also several defensive strategies.
In all cases, the proposed method outperforms reference techniques.


\section{Related Work}

Early query-based black-box attacks to deep neural networks appeared almost simultaneously in 2017.
In \cite{Chen2017zoo} a zeroth order stochastic (ZOO) coordinate descent algorithm is proposed,
based on the coordinate-wise ADAM optimizer or on coordinate-wise Newton’s method.
Both solutions modify dynamically the intensity of the attack, the former proving eventually more effective.
Several strategies are proposed to reduce the number of queries: dimension reduction via bilinear interpolation, hierarchical attack and importance sampling.

In \cite{Narodytska2017} a greedy local search is carried out aimed at perturbing only the patches that impact most on decision.
The adversarial samples, however, exhibit odd patches, which are easily spotted and are sort of a trademark of the attack.
Moreover, a very large number of queries is necessary to complete the attack, as observed also in \cite{Bhagoji2017exploring}.
Along the same line,
\cite{Su2019onepix} uses single-pixel adversarial perturbations, extended to groups of five pixels in \cite{Su2019de}.
The most suitable pixels to attack are selected through a simple genetic algorithm.
However, the success rate is rather low, below 80\%, and attacks are easily spotted and very fragile to any form of image processing.

\cite{Bhagoji2017exploring} uses a gradient estimation strategy followed by the classical FGSM attack, with random grouping to reduce the number of queries.
The method compares favourably with ZOO \cite{Chen2017zoo}, and proves effective also with real-world systems.
A further variation of ZOO, called AutoZOOM (autoencoder-based zeroth order optimization method) \cite{Tu2019}, uses an adaptive random gradient estimation strategy to balance query counts and distortion.
Complexity is reduced through subsampling, obtained by using an autoencoder trained off-line with unlabeled data or through bilinear interpolation.
Also \cite{Ilyas2018} pursues the main goal of limiting the number of queries.
To this end, it makes use of natural evolution strategies (NES) to estimate the black-box output gradient with respect to the attacked input image,
and generate suitable adversarial samples.
{\red NES is also adopted in \cite{Li2019nattack} with the aim of estimating the adversarial perturbation distribution of the network under attack,
so as to draw samples from such a distribution which are likely to fool the classifier.}

It is worth underlining that several elegant solutions, {\it e.g.,} \cite{Chen2017zoo,Tu2019,Ilyas2018}, rely on real-valued (that is, non-integer) perturbations.
The attack is iteratively refined based on closed-form optimization and estimation methods, and the changes are often quite subtle.
These small changes, however, are readily washed out at each query by any form of image compression or even just rounding, with a performance impairment,
as demonstrated experimentally in \cite{Duan2019}.
This is exactly what happens in all systems that accept queries only in integer-valued format,
an intrinsic form of defense \cite{Das2018}.
Hence, practical attacks should be designed and tested to prove robustness to such scenarios \cite{Marra2018}.

\section{Background}
\label{sec:background}
A CNN used for image classification computes a function, $y=f(x;\theta)$,
which maps an input image, $x \in \mathbb{R}^{n_1 \times n_2 \times n_3}$, into an output vector, $y \in \mathbb{R}^L$.
In the following, with no loss of generality, we will consider color images, hence $n_3=3$.
The specific function implemented by the CNN depends on the model parameters, $\theta$,
namely, the CNN weights which are learned during the training phase to optimize a suitable loss function.

Here, we consider two scenarios of practical interest.
In the first one, the classes are known in advance, and the system is asked to decide which class the query belongs to, and how reliable the decision is.
Accordingly, it provides in output a vector of probabilities
\begin{equation}
    y(i) = \Pr(x \in C_i), \hspace{3mm} i=1,\ldots,L
\end{equation}
also called confidence levels, with $C_i$ the $i$-th class, and $L$ equal to the number of classes.
The decision is made in favor of the maximum-probability class.
In the second scenario, a scalable system is considered, where the classes are not known in advance, and their number grows with time.
This applies, for example, to biometric identification systems, where new users keep being enrolled all the time.
In this case the CNN is used as a feature extractor.
For each input image, $x$, the CNN generates a discriminative vector of features, $y$, with length $L$ unrelated with the number of classes,
which is then used to perform the actual classification, for example, with a minimum distance rule.
In both cases, we assume the CNN to be already trained, with parameters $\theta$ defined once and for all.
Therefore, in the following we will neglect dependencies on $\theta$.

CNN-based systems are vulnerable to adversarial attacks.
To formalize the problem, we define:
$x_0$, the original image,
$y_0=f(x_0)$, the output vector associated with it,
$x^a=x_0+W$, the modified image, with $W$ the additive adversarial noise,
$y^a=f(x^a)$, the output vector associated with it,
and $y^t$, the target output vector.
Moreover, we introduce $\L(y_1,y_2)$, a suitable loss function measuring vector mismatch in the output domain, and $D(x_1,x_2)$,
a suitable measure of the image-domain distortion.
The attacker's aim is to generate a new image, $x^a$,
that is close to the original in the source domain, hence small $D(x^a,x_0)$,
but whose associated vector is close to the target vector in the output domain, small $\L(y^a,y^t)$,
as described pictorially in Fig.\ref{fig:mapping}.

We cast the problem as a constrained optimization, setting a limit on the acceptable image distortion, $D_{\max}$.
Accordingly, the attacker looks for the image $x^a$ defined by
\begin{equation}
    x^a = \arg\min_{x} \L(f(x),y^t), \hspace{3mm} \mbox{s.t. } D(x,x_0)\leq D_{\max}
\end{equation}

For typical classification problems, the loss of choice is the cross-entropy.
Instead, when the CNN is used for feature extraction, we will consider the Euclidean distance between the extracted feature vectors.

As for the attack strategies,
starting from a given image, $x$, the small perturbation $\Delta W$ that maximizes the loss decrease is, by definition,
proportional to the gradient of the loss itself with respect to $x$
\begin{equation}
    \Delta W \sim \nabla_{x} \L(f(x),y^t)
\end{equation}
If the loss is differentiable and the CNN is perfectly known with its parameters, namely, in the white box scenario,
the gradient can be computed through backpropagation, allowing very effective attack strategies.
In a black-box (BB) scenario, instead, the attacker has no information about the model architecture, its parameters, $\theta$, or the training set used to learn them.
However, the attacker can query the system at will, and use the corresponding outputs to estimate the gradient.
For any unit-norm direction of interest, $\phi$, the attacker collects the outputs, $y^+$ and $y^-$,
corresponding to the opposite queries $x^+=x+\epsilon\phi$ and $x^-=x-\epsilon\phi$, with $\epsilon$ suitably small,
and estimates the derivative of $\L$ along $\phi$ as,
\begin{equation}
    \frac{\partial\L}{\partial\phi} \simeq \frac{\L(y^+,y^t)-\L(y^-,y^t)}{2\epsilon}
\end{equation}
If the selected directions are the individual pixels, the whole gradient can be computed, but at the cost of a large number of queries.
Hence, practical algorithms approximate gradient-based BB attacks through more efficient strategies.

\section{Proposed method}
\label{sec:method}

We have the twofold goal to drive the classifier towards a desired decision {\em and} preserve a good image quality.
Now, in white-box systems, the gradient of the loss is known,
so perturbations are typically taken along the steepest descent path
and the effects on distortion are taken into account only a posteriori, by verifying the quality constraint and rejecting unsuitable perturbations.
In black-box systems, we cannot compute the gradient of the loss, if not by means of an inordinate number of queries.
However, we {\em can} compute the gradient of the distortion.
So, we follow the dual approach, and take perturbations along the path that increases distortion the least, verifying the loss reduction afterwards.

\begin{figure}
	\centering
	\renewcommand{\S}{1.6}
	\renewcommand{\tabcolsep}{2pt}
	\begin{tabular}{cccc}
		\includegraphics[scale=\S]{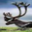} &
		\includegraphics[scale=\S]{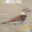} &
		\includegraphics[scale=\S]{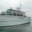} &
		\includegraphics[scale=\S]{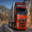} \\
		
		\includegraphics[scale=\S]{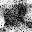} &
		\includegraphics[scale=\S]{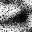} &
		\includegraphics[scale=\S]{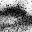} &
		\includegraphics[scale=\S]{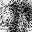} \\
		
		\includegraphics[scale=\S]{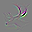} &
		\includegraphics[scale=\S]{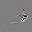} &
		\includegraphics[scale=\S]{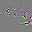} &
		\includegraphics[scale=\S]{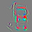} \\
		
		\includegraphics[scale=\S]{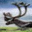} &
		\includegraphics[scale=\S]{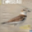} &
		\includegraphics[scale=\S]{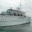} &
		\includegraphics[scale=\S]{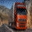}
	\end{tabular}
	\caption{From top to bottom: original CIFAR10 image, SSIM gradient, saliency map, attacked image.
    PQP works on low-SSIM gradient pixels, often image edges and keypoints, characterized by high saliency.
    The attacked image is very similar to the original and does not show unnatural artifacts.
	\label{fig:saliency}
    }
\end{figure}

With Fig.\ref{fig:saliency} we explain the rationale of our approach.
The first row shows some 32$\times$32-pixel images of the CIFAR10 dataset.
On the second row, the corresponding SSIM gradient images are shown.
Note that, contrary to what happens with $L_p$ distortion laws, perturbations cause a smaller or larger increase in distortion depending on the local context.
In active areas of the image, with keypoints, boundaries and textures, errors are less visible and, accordingly, the SSIM gradient is smaller, while it is larger in flat areas, where errors stand out.
Therefore, working in low-SSIM gradient areas, we ensure that attacks have low visibility.
On the other hand, it is exactly these features, not background areas, that the CNN takes most into account for classification.
This appears clearly in the saliency maps shown on the third row, obtained for the VGG16 net, where bright colors highlight pixels that contribute most to the final decision.
Therefore, the very same areas that ensure low quality gradient are also the most important for classification,
and SSIM-guided perturbations are not only inconspicuous but also very effective.
The fourth row shows the final (successful) adversarial samples produced by PQP, barely distinguishable from the originals and free from unnatural artifacts.
We note, in passing, that SSIM was already used in the literature \cite{Rozsa2016}, but only for white-box attacks, and only for checking the effect of perturbations, not for driving them.

Before moving to describe the proposed PQP method in detail, let us consider a naive algorithm which embodies its concepts in the simplest form.
At each step, this algorithm computes the gradient of the SSIM, $\G=\nabla_x \SSIM(x,x_0)$, with respect to the current attack, $x$,
and selects the component where a perturbation causes the least increase of the SSIM
\begin{equation}
    (s_1,c_1) = \arg\min_{s,c} |\G(s,c)|
\end{equation}
with $s$ the spatial coordinates and $c$ the color band.
Then, two new images $x^+=x+1_{s_1,c_1}$ and $x^-=x-1_{s_1,c_1}$ are generated which differ from $x$ by 1 level only at the selected component.
The system is probed with these queries, and the one that most reduces the loss becomes the new attacked image.
These steps are then repeated until convergence.
Note that the SSIM gradient can be computed with negligible cost both in closed form \cite{Avanaki2008SSIM}, and by means of a simple convolutional network, as we do here (see appendix).

\begin{algorithm}[!t]
\footnotesize
\begin{algorithmic}[1]
    \Require{$x_0, y^t, D_{\max}, \L_{\min}, Q, N, \delta, k_{\max}$}
    \Ensure{$x^a$}
    \State $x=x_0$
    \While{$D(x,x_0)<D_{\max}$ {\bf and} $\L(f(x),y^t)>\L_{\min}$}
        \State $\G=\nabla_{x} \SSIM(x,x_0)$
        \State $M_{\rm low} = {\rm Segment}(\G,Q)$                           \Comment{select low-gradient pixels}
        \State $k=0$
        \State $\Delta\L=0$
        \While{$k<k_{\max}$ {\bf and} $\Delta\L \geq 0$}
            \State $\Delta W = {\rm Perturbation}(M_{\rm low},N,\delta$)    \Comment{generate $\Delta W$}
            \State $\L^+ = \L(f(x+\Delta W),y^t)$                           \Comment{BB query}
            \State $\L^- = \L(f(x-\Delta W),y^t)$                           \Comment{BB query}
            \State $\Delta\L = \min(\L^+,\L^-)-\L(f(x),y^t)$
            \State $k=k+1$                                                  \Comment{if $k$=$k_{\max}$ accept anyway}
        \EndWhile
        \If{$\L^+<\L^-$}
            \State $x=x+\Delta W$
        \Else
            \State $x=x-\Delta W$
        \EndIf
    \EndWhile
    \State $x^a=x$;
\end{algorithmic}
\caption{PQP}
\label{algo:PQP}
\end{algorithm}

This naive algorithm, however, suffers from two practical problems:
{\it  i)} by modifying only one pixel at a time, and by only one level, it becomes exceedingly slow, and
{\it ii)} by choosing deterministically the pixel to modify, it is easily trapped in local minima.
The actual PQP method, described by the pseudo-code of Procedure \ref{algo:PQP}, overcomes both these problems.
In particular, the procedure of line 8 generates a perturbation, $\Delta W$, which modifies $N$ pixels at once and by $\pm\delta$ levels.
Suitable choices of $N$ and $\delta$ allow one to speed up the attack considerably with respect to pixel-wise perturbations, without a significant quality impairment.
Like in the naive algorithm, the net is queried twice, with $x+\Delta W$ and $x-\Delta W$, and the query which most reduces the loss is accepted.
If neither reduces the loss, a new random perturbation is generated, with the same rules.
After $k_{\max}$ unsuccessful attempts, the image is modified anyway to escape local minima.

To avoid local minima, instead, the $N$ pixels to modify are chosen randomly,
with the only constraint to belong to the set $M_{\rm low}$ comprising the $Q\%$ pixels with the lowest SSIM gradient.
Choosing $Q$ large enough, {\it e.g.} 50\% of the image, ensures high attack variety while avoiding any sharp quality degradation.

A third improvement consists in using {\em color-coherent} perturbations, such that the sign of perturbations is the same for all color components of a pixel.
This choice prevents abrupt, highly visible, variations of the hue, and proved experimentally to improve the attack.

\section{Experimental analysis}
\label{sec:experiments}

In this Section, we analyze the performance of the proposed black-box attack
both for general-purpose object recognition, with the popular CIFAR dataset, and for a biometric face recognition task, using the MCS2018 dataset.
As performance metrics we will consider
the success rate of the attack, SR, the number of black-box queries necessary to complete the attack, NQ, and the average distortion of successfully attacked images.
{\red
To measure distortion we consider not only SSIM, used itself to guide the attack, but also
two more full-reference measures, PSNR (peak signal-to-noise ratio), and VIF (visual information fidelity) \cite{VIF_sheikh2006image}, and
two no-reference ones, BRISQUE (blind/referenceless image spatial quality evaluator) \cite{BRISQUE_mittal2012no}, and PIQE (perception-based image quality evaluator) \cite{PIQE_venkatanath2015blind}.}
Note that number of queries and image quality are computed only on successfully attacked images.

We will compare results with several baselines and state-of-the-art references:
IFD (iterative finite differences) \cite{Bhagoji2017exploring} is the main baseline, with pixel-wise perturbations, slow but characterized by high success rate;
IGE-QR-RG (iterative gradient estimation with query reduction by random grouping) is a fast method proposed in \cite{Bhagoji2017exploring} based on group-wise perturbations;
{\red Local Search Attack (LSA) \cite{Narodytska2017} is a method that only perturbs the most salient pixels in the image;}
AutoZOOM-B is the version of AutoZOOM \cite{Tu2019} which uses bilinear interpolation to gain efficiency with no need of prior information;
NES-LQ is the limited-query method proposed in \cite{Ilyas2018} based on a natural evolution strategy (NES);
{\red \NA\ \cite{Li2019nattack} is another recently proposed NES-based approach that draws adversarial perturbations from their estimated distribution.}
In addition, we provide results also for the
white-box I-FGSM (iterative fast gradient sign method) attack \cite{Kurakin2016}, as a sort of upper bound for the performance of black-box methods.
We use the code published by the authors online \cite{Bhagoji2017code,Tu2018code,Ilyas2018code}, with the parameters suggested in the original papers, to which the reader is referred for any further detail.
For the proposed method, based on some preliminary experiments, analyzed in Tab.\ref{tab:PQP_parameters}, we set $Q$=66\% $N$=20, $\delta$=1, and $k_{\max}$=20.
Our own code is published online \cite{Gragnaniello2019code}.
Unless otherwise specified, we consider a 8-bit integer setting, with images rounded if necessary after each iteration.

\subsection{Object recognition with the CIFAR10 dataset}
\begin{figure}[t]
	\centering
	\includegraphics[width=0.42\textwidth]{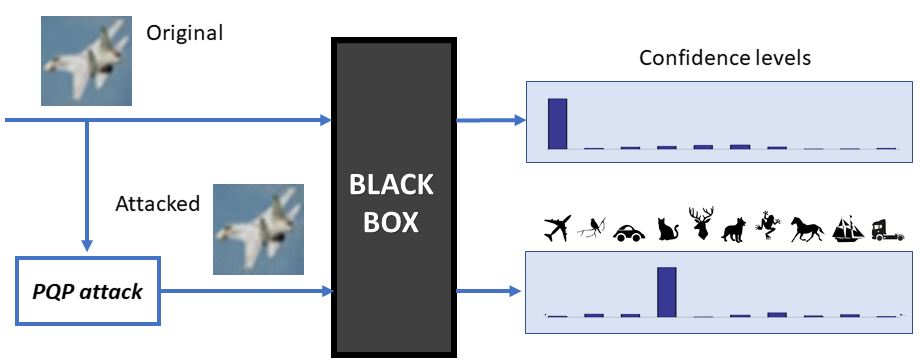}
	\caption{The CIFAR10 attack scenario. The attacker injects some adversarial noise to increase the confidence level of the target class.}
	\label{fig:CIFAR}
\end{figure}

The popular CIFAR10 object recognition dataset comprises 60000 32$\times$32-pixel RGB images, 50000 for training and 10000 for testing, equally distributed among 10 classes.
In the experiments, we consider the most challenging task and target the class with the lowest confidence score.
The attack stops either when the confidence of the selected target class exceeds 0.9, in which case we label the attack as successful,
or when the SSIM goes below 0.95, a case of unsuccessful attack.
Fig.\ref{fig:CIFAR} depicts the CIFAR10 attack scenario.

Tab.\ref{tab:cifar_resnet32} shows results for the ResNet32 \cite{He2016} classifier but similar results have been observed with other deep networks,
which are not reported here for brevity.
{\red To allow for an easier inspection of results, near each indicator an arrow indicates whether good performances correspond to large $\uparrow$ or small $\downarrow$ numbers.}
First of all, we observe a huge gap between the success rates of IFD, IGE-QR-RG, and PQP, all very close to 100\%, and those of LSA, AutoZOOM-B, and NES-LQ, much lower.
This is not surprising.
While the former are designed or can be adapted to work in our integer-valued setting,
the latter are intrinsically real-valued, with perturbations that are largely washed out by rounding, leading to mostly ineffective attacks,
despite all attempts to optimize parameters for this scenario.
With such low SRs, of course, it makes no sense to compute the other performance indicators.
All integer-based methods attack quite easily this classifier, ensuring also a very low distortion, with a minimum SSIM of 0.985.
With such high success rates and good quality indicators, comparable to those of the white-box attack, the truly discriminative metric is the number of queries.
Under this point of view, the proposed method is about 3 times faster than IGE-QR-RG, and 15 times faster than the pixel-wise baseline (a comparison with the white-box attack makes no sense).
{\red As for \NA, with over 10000 queries on the average, it turns out to be rather inefficient, contrary to what observed in \cite{Li2019nattack}.
This is very likely due to our more challenging scenario, with worst-case targeted attacks, which renders impractical to estimate the distribution of adversarial noise.
In summary, PQP ensures successful attacks with a small number of queries.
Moreover, no quality loss is observed, not only in terms of SSIM but of all quality measures, even with a significant improvement in terms of the no-reference PIQE.}

\renewcommand{\tabcolsep}{3pt}
\begin{table}[!t]
\caption{Attacking ResNet32 on CIFAR10.}
\centering
\footnotesize
\begin{tabular}{l|r|r|rrrrr}
	\toprule
	\ru Attack      &      SR\U~ &     NQ\D~ &    SSIM\U &    PSNR\U & \red VIF\U & \red BRI.\D & \red PIQE\D \\ \midrule\midrule
	\ru I-FGSM (WB) &      98.69 &         9 &     0.991 &     42.65 &  \red10.51 &   \red39.09 &   \red38.92 \\ \midrule
	\ru IFD         &      99.01 &     26639 &     0.991 &     42.67 &  \red10.53 &   \red39.10 &   \red39.08 \\
	\ru IGE-QR-RG   &      95.84 &      5467 &     0.985 &     40.84 &   \red9.54 &   \red39.00 &   \red38.92 \\
	\ru \red LSA    & \red ~0.00 &   \red -- &   \red -- &   \red -- &    \red -- &     \red -- &     \red -- \\
	\ru AutoZOOM-B  &      ~0.00 &        -- &        -- &        -- &         -- &          -- &          -- \\
	\ru NES-LQ      &      ~8.74 &        -- &        -- &        -- &         -- &          -- &          -- \\
	\ru \red\NA     &  \red92.97 & \red11459 & \red0.986 & \red37.30 &   \red7.67 &   \red39.13 &   \red42.04 \\
	\ru PQP         &      98.61 &      1593 &     0.989 &     39.77 &   \red8.52 &   \red39.23 &   \red35.42 \\ \bottomrule
\end{tabular}
\label{tab:cifar_resnet32}
\end{table}

\renewcommand{\tabcolsep}{3pt}
\begin{table}[t]
\caption{Attacking ResNet32 with no rounding nor SSIM constraint.}
\centering
\footnotesize
\begin{tabular}{l|r|r|rrrrr}
	\toprule
	\ru Attack      &      SR\U~ &     NQ\D~ &    SSIM\U &    PSNR\U &     VIF\U &       BRI.\D &      PIQE\D \\ \midrule\midrule
	\ru \red LSA    & \red 20.44 &    \red-- &    \red-- &    \red-- &    \red-- &       \red-- &      \red-- \\
	\ru AutoZOOM-B  &      20.54 &        -- &        -- &        -- &        -- &           -- &          -- \\
	\ru NES-LQ      &     100.00 &      1495 &     0.880 &     27.56 &  \red4.36 &    \red41.16 &   \red37.73 \\
	\ru PQP         &      98.61 &      1585 &     0.990 &     39.82 &  \red8.45 &    \red39.24 &   \red35.37 \\ \bottomrule
\end{tabular}
\label{tab:cifar_LQ}
\end{table}

\renewcommand{\tabcolsep}{3pt}
\begin{table}[t]
\caption{Attacking ResNet32 on CIFAR10 with NN defense.}
\centering
\footnotesize
\begin{tabular}{l|r|r|rrrrr}
	\toprule
	\ru Attack      &      SR\U~ &     NQ\D~ &    SSIM\U &    PSNR\U &     VIF\U &       BRI.\D &      PIQE\D \\ \midrule\midrule
	\ru I-FGSM (WB) &      92.03 &        23 &     0.984 &     37.07 &      7.73 &        39.18 &       35.99 \\ \midrule
	\ru IFD         &      92.37 &     71420 &     0.984 &     37.09 &      7.73 &        39.23 &       35.99 \\
	\ru IGE-QR-RG   &      81.37 &     11278 &     0.980 &     36.04 &      7.31 &        39.49 &       36.79 \\
	\ru \red\NA     &  \red84.00 &     38419 & \red0.977 & \red34.65 &  \red6.91 &    \red39.88 &   \red39.99 \\
	\ru PQP         &      95.81 &      3991 &     0.987 &     36.07 &      6.61 &        39.16 &       34.57 \\ \bottomrule
\end{tabular}
\label{tab:cifar_resnet32_triplets}
\end{table}

The above experiment shows that some inherently ``floating-point'' attacks, LSA, AutoZOOM, and NES-LQ, do not fit well our integer-valued scenario, and hence will not be considered anymore in our analysis.
However, to establish a fair comparison with these state-of-the-art methods,
we carry out a further experiment on the CIFAR10 dataset, relaxing both the 8-bit integer constraint and the SSIM contraint.
Instead, we keep targeting the worst class and requiring a confidence level beyond 0.9 to declare a success.
Results are reported in Tab.\ref{tab:cifar_LQ}.
{\red
Even in this favourable scenario,
AutoZOOM-B and LSA keep providing poor results, with a success rate around 20\%.
NES-LQ, instead, ensures an excellent success rate, but the image quality shows a serious impairment uniformly for all measures, with a PSNR loss of about 10 dB, and a SSIM decrease of about 0.1 with respect to PQP.}
Overall, these results suggest that the proposed algorithm works much better than reference methods not only in our challenging ``real-world'' scenario,
but also in a more abstract scenario in which all constraints on data format are removed.

To complete our realistic analysis on the CIFAR10 dataset, we consider the presence of countermeasures.
Indeed, with the diffusion of adversarial attacks, some defense strategies are becoming widespread, like the so-called NN (nearest neighbor) defense.
For each test image, $x$, the net extracts a feature vector, called $F$ here,
computes its distance from the centroid of all classes, $\overline{F}_1,\ldots,\overline{F}_{10}$, and decides in favor of the closest one.
The NN classifier is somewhat inferior to the standard one but more robust to adversarial attacks, and fully sensible for biometric authentication systems.
Here, we consider NN defense with the worst-case attack, where the target class is $t=\arg\max_c \|F-\overline{F}_c\|_2$.
The attack is successful if the feature vector of the attacked image, $F^a$, becomes closer to $\overline{F}_t$ than to all other centroids,
and the distance $\|F^a-\overline{F}_t\|_2$ goes below a given threshold $\gamma$.
We set $\gamma=0.2$, which is the average distance between the features of a class and the corresponding centroid,
while the average distance between the test images and the centroid of their target-class (distance before attack) is 0.81.
Results reported in Tab.\ref{tab:cifar_resnet32_triplets} show that the new classifier is more robust to attacks.
Success rates reduce significantly, even for the white-box reference, going down to 81\% for IGE-QR-RG and to 84\% for \NA.
PQP, instead, keeps ensuring a good success rate, the smallest number of queries among black-box methods, and a very good image quality.

\subsection{Face recognition with the MCS2018 dataset}

\begin{figure}[!t]
	\centering
	\includegraphics[width=0.42\textwidth]{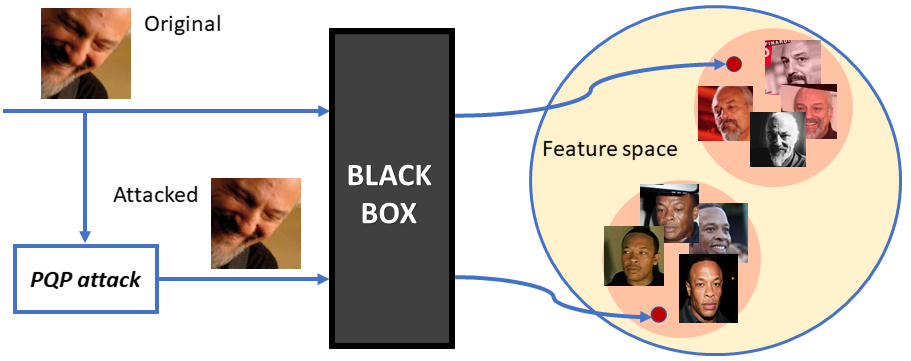}
	\caption{The MCS2018 attack scenario (implicit NN defense). The injected adversarial noise brings the output feature vector close to the feature vectors associated with the sample images of the target subject.}	
\label{fig:MCS2018}
\end{figure}

Evaluating the robustness of a face recognition system is of primary importance to improve
its performance in an adversarial setting.
This task was originally proposed in the {\em ``Adversarial Attacks on Black Box Face Recognition''} competition, in the context of the MachinesCanSee conference (MCS2018) held in Moscow in June 2018.
The reference scenario is depicted in Fig.\ref{fig:MCS2018}.
A black-box face recognition system is provided,
which extracts a 512-component unit-norm feature vector from each submitted 112$\times$112 RGB image.
This vector is then fed to a database to single out the best matching identity.
In the competition, the attacker was asked to imperceptibly modify an input image associated with a given identity,
with the aim of tricking the system into recognizing a different specific identity.
1000 pairs of source-target identities $(S_i,T_i)$ were provided, with 5 images for each identity: $S_i \to \{x^s_{i1},\ldots,x^s_{i5}\}$, and $T_i \to \{x^t_{i1},\ldots,x^t_{i5}\}$.
The goal of the attacker was to move the feature vector of the attacked image $F^a_{ij}$ close to the centroid, $\overline{G}_i$, of the target identity features,
subject to ${\rm SSIM}(x^s_{ij},x^{s,a}_{ij})\geq 0.95$.
In the competition, we (the GRIP team) obtained eventually the lowest average distance \cite{CODALABresults}, $D$=0.928,
using a preliminary version of the proposed algorithm.

\renewcommand{\tabcolsep}{8pt}
\begin{table}[!t]
\renewcommand{\d}{$^\ast$}
\centering
\footnotesize
\caption{Choosing parameters on the MCS2018 dataset ($\gamma=1.0$).}
\begin{tabular}{ccc||r|r|rr}	
    \toprule
    \ru   $Q$ & $N$ & $\delta$ &   SR\U~ &   NQ\D~ &  SSIM\U &  PSNR\U \\ \midrule\midrule
    \ru  66\% &  20 &        1 &   94.20 &   13265 &   0.976 &   38.62 \\ \midrule
    \ru  33\% &  -- &       -- &   98.00 &   18348 &   0.981 &   37.74 \\
    \ru 100\% &  -- &       -- &   72.00 & \d10636 & \d0.968 & \d39.41 \\ \midrule
    \ru    -- &  10 &       -- &   97.60 &   20093 &   0.981 &   39.55 \\
    \ru    -- &  40 &       -- &   85.00 &    8507 &   0.972 &   37.69 \\ \midrule
    \ru    -- &  -- &        2 &   63.20 &  \d5163 & \d0.966 & \d36.91 \\
    \ru    -- &  -- &        3 &   27.80 &      -- &      -- &      -- \\ \bottomrule
\multicolumn{7}{l}{\ru Results marked with ``*'' may be overly optimistic.}
\end{tabular}
\label{tab:PQP_parameters}
\end{table}

Here, we modify slightly the attacker's goal, in order to define a success rate and allow for a meaningful comparison with the other tasks.
We declare a success for image $x^s_{ij}$ when $\|F^a_{ij}-\overline{G}_i\|<\gamma$, with $\gamma$ a suitably chosen threshold, a failure when the SSIM goes below 0.95.
Since the feature vectors have unitary norm, unrelated vectors are nearly orthogonal, with a distance close to $\sqrt{2} \simeq 1.414$.
Features associated to the same identity are much closer, with average intra-class distance 0.903.
Therefore, we consider two cases, $\gamma=1.0$ and $\gamma=0.9$, the latter being the most challenging.

To select the PQP's parameters used in all experiments, $Q$=66\%, $N$=20 and $\delta$=1, we carried out some preliminary tests on the MCS2018 dataset.
Tab.\ref{tab:PQP_parameters} shows, for $\gamma$=1, the effect of perturbing these parameters with respect to their default values.
Note that statistics are computed only on the successful attacks, hence they are overly optimistic when \mbox{SR $\ll$ 100\%}, in which case we mark results with an asterisk.
These tests prove clearly the importance of working in the low-SSIM gradient region, as blind attacks ($Q$=100\%) cause a sharp decrease of both success rate and quality.
Moreover, they speak against increasing $\delta$ to speed up the attack, while increasing $N$ is less detrimental to quality.

\begin{figure*}
\newcommand{\WW}{15mm}  
\newcommand{\CA}{70}  
\newcommand{\CB}{50} 
\newcommand{\CC}{5} 
\newcommand{\CD}{25}  
\newcommand{\CAA}{50}  
\newcommand{\CBA}{60} 
\newcommand{\CCA}{35} 
\newcommand{\CDA}{25}  
\small
\centering
\begin{tabular}{c@{\hspace{1.5mm}}||@{\hspace{1.5mm}}c@{\hspace{1.5mm}}c@{\hspace{1.5mm}}c@{\hspace{1.5mm}}c@{\hspace{1.5mm}}c@{\hspace{1.5mm}}c@{\hspace{1.5mm}}c@{\hspace{1.5mm}}c@{\hspace{1.5mm}}c@{\hspace{1.5mm}}c}
	\includegraphics[width=\WW]{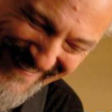} &
	\includegraphics[width=\WW]{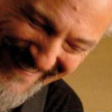} &
	\includegraphics[width=\WW]{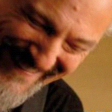} &
	\includegraphics[width=\WW]{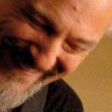} &
	\includegraphics[width=\WW]{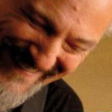} &
	\includegraphics[width=\WW]{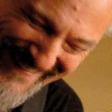} &
	\includegraphics[width=\WW]{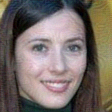} &
	\includegraphics[width=\WW]{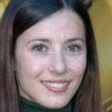} &
	\includegraphics[width=\WW]{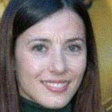} &
	\includegraphics[width=\WW]{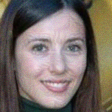} &
	\includegraphics[width=\WW]{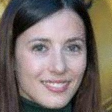} \\
	
	\includegraphics[width=\WW, trim={\CA, \CB, \CC, \CD}, clip]{sub1_or.png} &
	\includegraphics[width=\WW, trim={\CA, \CB, \CC, \CD}, clip]{sub1_I-FGSM_1_0.png} &
	\includegraphics[width=\WW, trim={\CA, \CB, \CC, \CD}, clip]{sub1_Pixel-based_1_0.png} &
	\includegraphics[width=\WW, trim={\CA, \CB, \CC, \CD}, clip]{sub1_RS_1_0.png} &
	\includegraphics[width=\WW, trim={\CA, \CB, \CC, \CD}, clip]{sub1_Nattack_1_0.png} &
	\includegraphics[width=\WW, trim={\CA, \CB, \CC, \CD}, clip]{sub1_PQP_1_0.png} &
	\includegraphics[width=\WW, trim={\CAA, \CBA, \CCA, \CDA}, clip]{sub3_I-FGSM_1_0.png} &
	\includegraphics[width=\WW, trim={\CAA, \CBA, \CCA, \CDA}, clip]{sub3_Pixel-based_1_0.png} &
	\includegraphics[width=\WW, trim={\CAA, \CBA, \CCA, \CDA}, clip]{sub3_RS_1_0.png} &
	\includegraphics[width=\WW, trim={\CAA, \CBA, \CCA, \CDA}, clip]{sub3_Nattack_1_0.png} &
	\includegraphics[width=\WW, trim={\CAA, \CBA, \CCA, \CDA}, clip]{sub3_PQP_1_0.png} \\
	
	\includegraphics[width=\WW]{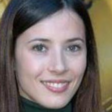} &
	\includegraphics[width=\WW]{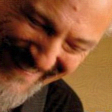} &
	\includegraphics[width=\WW]{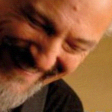} &
	\includegraphics[width=\WW]{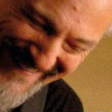} &
	\includegraphics[width=\WW]{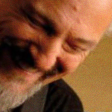} &
	\includegraphics[width=\WW]{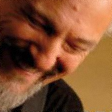} &
	\includegraphics[width=\WW]{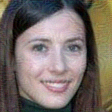} &
	\includegraphics[width=\WW]{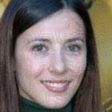} &
	\includegraphics[width=\WW]{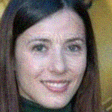} &
	\includegraphics[width=\WW]{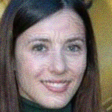} &
	\includegraphics[width=\WW]{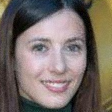} \\
	
	\includegraphics[width=\WW, trim={\CAA, \CBA, \CCA, \CDA}, clip]{sub3_or.png} &
	\includegraphics[width=\WW, trim={\CA, \CB, \CC, \CD}, clip]{sub1_I-FGSM_0_9.png} &
	\includegraphics[width=\WW, trim={\CA, \CB, \CC, \CD}, clip]{sub1_Pixel-based_0_9.png} &
	\includegraphics[width=\WW, trim={\CA, \CB, \CC, \CD}, clip]{sub1_RS_0_9.png} &
	\includegraphics[width=\WW, trim={\CA, \CB, \CC, \CD}, clip]{sub1_Nattack_0_9.png} &
	\includegraphics[width=\WW, trim={\CA, \CB, \CC, \CD}, clip]{sub1_PQP_0_9.png} &
	\includegraphics[width=\WW, trim={\CAA, \CBA, \CCA, \CDA}, clip]{sub3_I-FGSM_0_9.png} &
	\includegraphics[width=\WW, trim={\CAA, \CBA, \CCA, \CDA}, clip]{sub3_Pixel-based_0_9.png} &
	\includegraphics[width=\WW, trim={\CAA, \CBA, \CCA, \CDA}, clip]{sub3_RS_0_9.png} &
	\includegraphics[width=\WW, trim={\CAA, \CBA, \CCA, \CDA}, clip]{sub3_Nattack_0_9.png} &
	\includegraphics[width=\WW, trim={\CAA, \CBA, \CCA, \CDA}, clip]{sub3_PQP_0_9.png} \\
\multicolumn{1}{c}{Original} & I-FGSM & IFD & {\footnotesize IGE-QR-RG} & \red\NA & PQP & I-FGSM & IFD & {\footnotesize IGE-QR-RG} & \red\NA & PQP
\end{tabular}
	\caption{\red Example attacks on two MCS2018 images with $\gamma=1.0$ (top) and $\gamma=0.9$ (bottom). Originals are shown in the left column.
    For the ``bearded man'', a light adversarial noise is sufficient, and attacked images show little signs of distortion, more visible in the flat areas except for PQP.
    A heavier adversarial noise is necessary to attack the ``girl'',  with visible distortions and many ``weird'' patters. PQP images are also distorted, but keep a natural appearance.}
	\label{fig:sample_attacks}
\end{figure*}

Tab.\ref{tab:MCS2018_SSIM} shows results for attacks to the MCS2018 face recognition system
(to carry out experiments in a reasonable time, we implemented a replica of the competition system based on the organizer's description and software, and testing strict compliance of results).
Given the stringent MCS2018 rules, attacks are much more difficult than in the CIFAR10 case.
Even with the larger distance threshold, $\gamma$=1.0, a level which does not always ensure reliable face identification, the IGE-QR-RG attack fails most of the times{\red, while \NA\ only reaches 64\% success rate}.
At the same level, however, the proposed PQP has a success rate beyond 95\%, even better than I-FGSM and IFD.
With $\gamma$=0.9 all performance metrics worsen, of course, and only PQP keeps ensuring a success rate over 80\%.
It is also worth underlining the large improvement with respect to PQP-$\alpha$, the preliminary version which ranked first in the MCS2018 competition, called ``grip'' in that context.
{\red
As for image quality, it is guaranteed in all cases by the 0.95 constraint on the SSIM, and PQP provides always the largest SSIM.
Nonetheless, turning to the other quality measures, results are more heterogeneous.
The \NA, in particular, seems to ensure a better image quality than PQP according to the VIF and BRISQUE measures, but much worse in terms of PIQE.
These contrasting indications are in part due to the fact that results are computed on a limited and biased sample (only successful attacks).
Another warning concerns no-reference quality estimators, which are typically trained on large patches drawn from high-resolution images, whose statistics largely differ from those used in our experiments.
All these facts suggest to take these data with a grain of salt, and refer ultimately to visual inspection of attacked images, as in Fig.\ref{fig:sample_attacks}.}

In order to gain a better insight into the performance of all methods, we consider a ``lighter'' version of the MCS2018 system, which simulates a less watchful human visual inspection.
Accordingly, the constraint on the SSIM is replaced by a much weaker constraint on PSNR, required only to exceed 30 dB.
With these rules, success rates are always very high, never less than 90\%, allowing for the collection of significant performance data.
{\red
In these conditions, PQP is about as fast as IGE-QR-RG, but ensures a much better quality according to all measures except BRISQUE.
Also the \NA\ exhibits better VIF and BRISQUE numbers but much worse SSIM and PIQE numbers and, more important, it requires a much larger number of queries.
Turning to the most interesting case of $\gamma$=0.9,
PQP keeps providing an average SSIM above 0.95, while for IGE-QR-RG this drops well below 0.9, which entails very likely visible distortions.
The \NA\ keeps exhibiting contrasting image quality results, but now with a much smaller success rate.
In general, we believe SSIM and PSNR to be the most reliable indicators of quality, at least for our scenario characterized by very small images subject to adversarial attacks.
}

\renewcommand{\tabcolsep}{2pt}
\renewcommand{\d}{$^\ast$}
\begin{table}[!t]
\centering
\footnotesize
\caption{Attacking the MCS2018 face recognition system.}
\begin{tabular}{l|l|r|r|rrrrr}
	\toprule
	$\gamma$             & \ru Attack               &     SR\U~ &       NQ\D~ &      SSIM\U &      PSNR\U &      VIF\U &       BRI.\D &      PIQE\D \\ \midrule\midrule
	\multirow{6}{*}{1.0} & \ru I-FGSM (WB)          &     84.62 &           7 &       0.975 &       40.40 &       5.65 &        30.94 &       23.69 \\
	\cmidrule{2-9}       & \ru IFD                  &     81.74 &      454743 &       0.971 &       39.84 &       7.14 &        30.80 &       23.41 \\
	                     & \ru IGE-QR-RG            &     16.44 &          -- &          -- &          -- &         -- &           -- &          -- \\
	                     & \ru \red\NA              & \red64.08 & \red\d17798 & \red\d0.970 & \red\d38.23 & \red\d8.28 &  \red\d31.73 & \red\d27.57 \\
	                     & \ru PQP-$\alpha$ \winner &     66.68 &     \d11883 &     \d0.965 &     \d34.46 &     \d6.30 &      \d29.53 &     \d13.36 \\
	                     & \ru PQP                  &     95.80 &       13400 &       0.978 &       37.13 &       6.34 &        34.66 &       14.69 \\ \midrule\midrule
	\multirow{6}{*}{0.9} & \ru I-FGSM (WB)          &     62.76 &         \d9 &     \d0.970 &     \d39.25 &     \d5.80 &      \d31.46 &     \d24.20 \\
	\cmidrule{2-9}       & \ru IFD                  &     56.86 &    \d536919 &     \d0.967 &     \d38.93 &     \d7.58 &      \d31.48 &     \d25.22 \\
	                     & \ru IGE-QR-RG            &      4.66 &          -- &          -- &          -- &         -- &           -- &          -- \\
	                     & \ru \red\NA              & \red35.20 & \red\d24066 & \red\d0.966 & \red\d37.36 & \red\d7.92 & \red \d32.09 & \red\d25.78 \\
	                     & \ru PQP-$\alpha$ \winner &     39.88 &     \d14550 &     \d0.962 &     \d33.87 &     \d6.44 &      \d30.39 &     \d13.44 \\
	                     & \ru PQP                  &     83.52 &       18244 &       0.973 &       35.90 &       6.61 &        34.11 &       15.24 \\ \bottomrule
	\multicolumn{9}{l}{\ru Results marked with ``*'' may be overly optimistic.}
\end{tabular}
\label{tab:MCS2018_SSIM}
\end{table}

\renewcommand{\tabcolsep}{2pt}
\begin{table}[!t]
\centering
\footnotesize
\caption{Attacking a ``lighter'' MCS2018 face recognition system.}
\begin{tabular}{l|l|r|r|rrrrr}
	\toprule
	$\gamma$             & \ru Attack               &     SR\U~ &       NQ\D~ &      SSIM\U &      PSNR\U &      VIF\U &       BRI.\D &      PIQE\D \\ \midrule\midrule
	\multirow{6}{*}{1.0} & \ru I-FGSM (WB)          &    100.00 &           9 &       0.968 &       39.71 &       5.45 &        30.21 &       23.16 \\
	\cmidrule{2-9}       & \ru IFD                  &    100.00 &      544140 &       0.964 &       39.18 &       6.67 &        30.28 &       22.34 \\
	                     & \ru IGE-QR-RG            &     99.98 &       17498 &       0.912 &       35.02 &       5.81 &        29.66 &       16.97 \\
	                     & \ru \red\NA              & \red99.12 &   \red25503 &   \red0.954 &   \red37.16 &   \red7.54 &    \red30.52 &   \red25.32 \\
	                     & \ru PQP-$\alpha$ \winner &     99.86 &       21286 &       0.954 &       33.90 &       5.97 &        27.76 &       13.36 \\
	                     & \ru PQP                  &    100.00 &       14620 &       0.977 &       36.98 &       6.19 &        34.97 &       14.71 \\ \midrule\midrule
	\multirow{6}{*}{0.9} & \ru I-FGSM (WB)          &     98.98 &          15 &       0.949 &       37.63 &       5.45 &        30.17 &       23.18 \\
	\cmidrule{2-9}       & \ru IFD                  &     99.16 &      893743 &       0.944 &       37.31 &       6.75 &        30.28 &       22.39 \\
	                     & \ru IGE-QR-RG            &     98.36 &       27410 &       0.878 &       33.56 &       5.79 &        29.70 &       17.16 \\
	                     & \ru \red\NA              & \red90.90 &   \red39728 &   \red0.940 &   \red35.74 &   \red6.70 &    \red30.40 &   \red22.48 \\
	                     & \ru PQP-$\alpha$ \winner &     97.20 &       59440 &       0.939 &       33.05 &       5.97 &        27.70 &       13.44 \\
	                     & \ru PQP                  &     99.60 &       30967 &       0.965 &       35.39 &       6.19 &        34.98 &       14.79 \\ \bottomrule
\end{tabular}
\label{tab:MCS2018_PSNR}
\end{table}

This is confirmed by Fig.\ref{fig:sample_attacks}, showing visual results for the various types of attacks on two MCS2018 images.
Apparently, the ``bearded man'' image is relatively simple to attack, and all methods introduce only limited distortion.
Nonetheless, for $\gamma$=0.9, weird geometrical patterns and color distortions appear in the flat areas of images attacked by all methods except PQP.
The ``girl'' image is obviously more difficult to attack, and visible distortions arise even for $\gamma$=1.
Also in this case, however, the PQP images appear more natural than the others, showing a lower quality than the original but without heavy patterns and color distortions.

\begin{figure}[!t]
	\centering
	{\color{white} .} \hspace{4mm} \includegraphics[width=0.9\columnwidth]{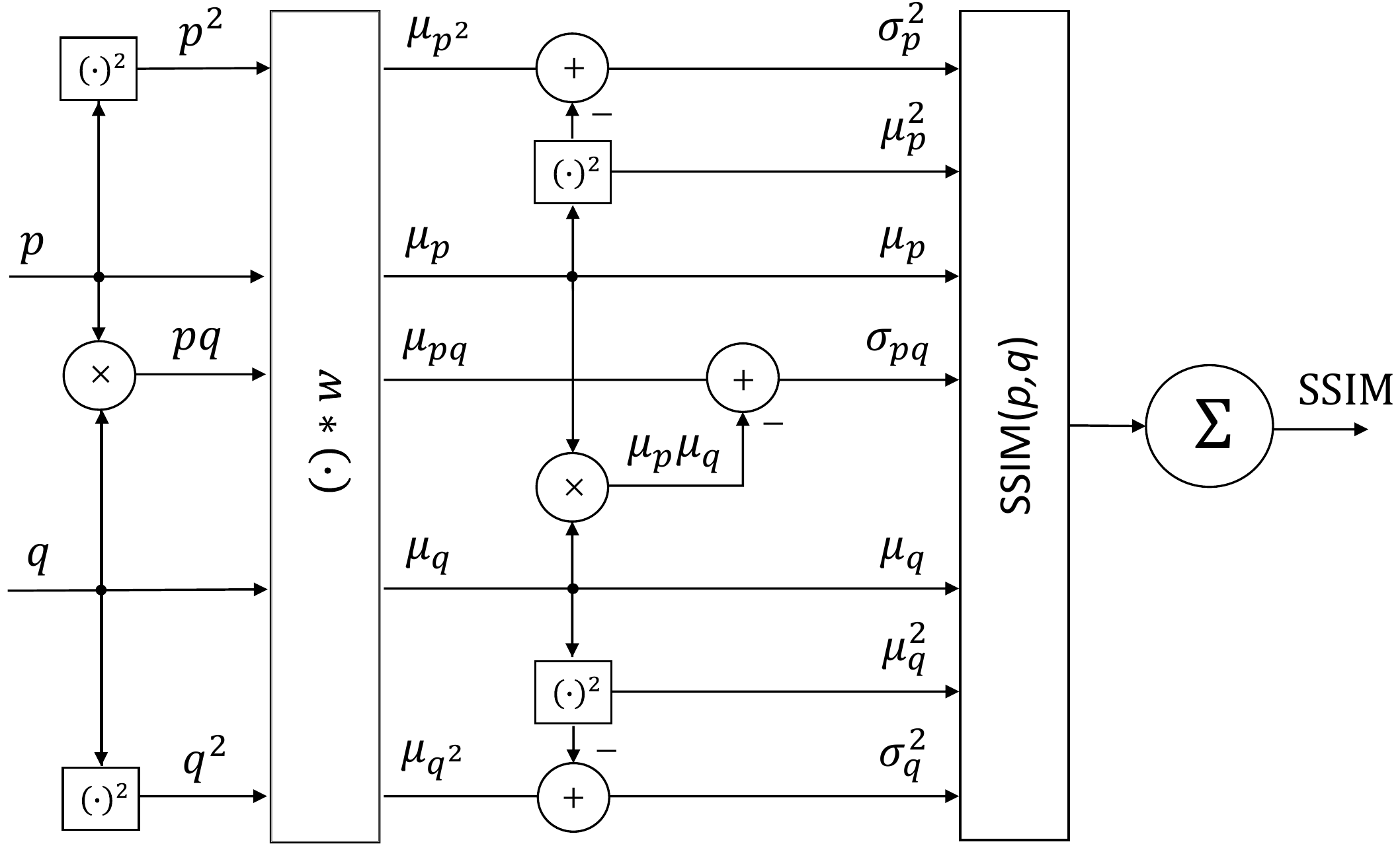}
	\caption{Computing SSIM gradient by back-propagation on a {\it ad hoc} CNN.}
	\label{fig:SSIMnet}
\end{figure}

\section{Conclusions}
We have proposed a new black-box method for attacking deep learning-based classifiers.
The image under attack is iteratively modified until the desired classification and confidence level are achieved.
However, changes are constrained to the low-SSIM gradient part of the image,
where perturbations have a limited impact on perceptual quality and, often, high impact on classification.
As a result, effective adversarial samples with high perceptual quality are rapidly generated.
This makes the proposed approach suitable to attack semi-automatic biometric authentication systems, usually implemented in surveillance systems of smart cities.

Experiments carried out in two quite different scenarios testify on the potential of the proposed approach.
Results are always very good, both with 8-bit integer and real-valued images, and do not impair much when some defense strategies are enacted.
The perceptual quality is always good,
visible distortions appear only in very small images, while adversarial noise introduced in regular-size images is always inconspicuous.


\section*{Appendix: Computing SSIM gradient by an {\it ad hoc} CNN}

\renewcommand{\SS}{{\rm SSIM}}
The SSIM between two homologous image patches $p$ and $q$ reads as
\begin{equation}
     \SS(p,q) = \frac{(2\mu_p\mu_q+\epsilon_1)(2\sigma_{pq}+\epsilon_2)}{(\mu^2_p+\mu^2_q+\epsilon_1)(\sigma^2_p+\sigma^2_q+\epsilon_2)}
\label{eq:patch_SSIM}
\end{equation}
Moments are computed with a suitable weighting window $w$ as in \cite{Wang2004}
\begin{equation}
\begin{split}
    \mu_{q}     = & \sum_{s \in \Omega} w(s)q(s) \\
    \sigma^2_q  = & \sum_{s \in \Omega} w(s)(q(s)-\mu_q)^2 \\
    \sigma_{pq} = & \sum_{s \in \Omega} w(s)(p(s)-\mu_p)(q(s)-\mu_q)
\end{split}
\label{eq:moments}
\end{equation}
with $s$ spanning the patch coordinates $\Omega$.
Similar formulas hold for $\mu_p, \sigma^2_p$.

To compute such weighted moments we use the simple CNN shown in Fig.\ref{fig:SSIMnet}, comprising a single convolutional layer and some further block for algebraic operations.
Hence, the gradient of the SSIM with respect to the whole image is then obtained by the usual back-propagation of the loss.

\bibliographystyle{model1-num-names}
\bibliography{refs}

\end{document}